\documentclass[sigconf]{acmart}
\AtBeginDocument{%
  }

\setcopyright{acmlicensed}
\copyrightyear{2018}
\acmYear{2018}
\acmDOI{XXXXXXX.XXXXXXX}

\acmConference[Conference acronym 'XX]{Make sure to enter the correct
  conference title from your rights confirmation emai}{June 03--05,
  2018}{Woodstock, NY}
\acmISBN{978-1-4503-XXXX-X/18/06}

\usepackage{graphicx}
\usepackage{textcomp}
\usepackage{xcolor}
\usepackage{bm}
\usepackage{soul}
\usepackage{flushend}
\usepackage{xspace}
\usepackage{soul}
\usepackage{algorithm}
\usepackage{algpseudocode}
\usepackage {diagbox}
\usepackage{url}
\def\system{\textit{CC-Net}\xspace}

\begin{document}

\title{Privacy-Preserved Taxi Demand Prediction System Utilizing Distributed Data}

\author{Ren Ozeki}
\affiliation{%
  \institution{Osaka University}
  \country{Japan}
}
\email{r-ozeki@ist.osaka-u.ac.jp}

\author{Haruki Yonekura}
\affiliation{%
  \institution{Osaka University}
  \country{Japan}
}
\email{h-yonekura@ist.osaka-u.ac.jp}

\author{Hamada Rizk}
\affiliation{%
  \institution{Tanta University, Egypt}
  \country{Osaka University,  Japan}
}
\email{hamada\_rizk@ist.osaka-u.ac.jp}

\author{Hirozumi Yamaguchi}
\affiliation{%
  \institution{Osaka University}
  \country{Japan}
}
\email{h-yamagu@ist.osaka-u.ac.jp}

\begin{abstract}
Accurate taxi-demand prediction is essential for optimizing taxi operations and enhancing urban transportation services. However, using customers' data in these systems raises significant privacy and security concerns. Traditional federated learning addresses some privacy issues by enabling model training without direct data exchange but often struggles with accuracy due to varying data distributions across different regions or service providers.
In this paper, we propose \system: a novel approach using collaborative learning enhanced with contrastive learning for taxi-demand prediction. Our method ensures high performance by enabling multiple parties to collaboratively train a demand-prediction model through hierarchical federated learning. In this approach, similar parties are clustered together, and federated learning is applied within each cluster. The similarity is defined without data exchange, ensuring privacy and security.
We evaluated our approach using real-world data from five taxi service providers in Japan over fourteen months. The results demonstrate that \system maintains the privacy of customers' data while improving prediction accuracy by at least 2.2\% compared to existing techniques.

\end{abstract}


\maketitle

\section{Introduction}
Using spatio-temporal location data can enhance various services, especially data-driven transportation, urban planning, and emergency management approaches\cite{sanchez2023prospects, 9172100, Yao_Wu_Ke_Tang_Jia_Lu_Gong_Ye_Li_2018, ozeki2024decentralized}.
One critical service, taxi transportation, is a key part of modern urban systems, providing convenient and efficient travel for many passengers.
However, there's often a mismatch between taxi supply and passenger demand, leading to higher cruising times, increased fuel consumption, and longer wait times, which reduces profits for taxi providers\cite{LIU2022100075, Ride-hailing-Worldwide:online}.

Taxi-demand prediction systems use data-driven approaches to forecast demand and optimize dispatch to address this\cite{Dynamic_Taxi_Ride-Sharing, Understanding_Taxi_Service_Strategies}.
Machine learning models trained on customer mobility data—including pickup and drop-off locations, routes, and timing—can predict future taxi demand in specific areas.
However, sharing this trajectory data raises significant privacy concerns, as it can reveal personal details like individuals' whereabouts and movement patterns\cite{may2020marc, schestakov2024re}.
Additionally, since the machine learning model that is overfitted with local data retains private information, privacy information can be leaked from the machine learning model by threats such as membership inference attacks\cite{pyrgelis2017knock, shokri2017membership}.

Differential privacy-based methods\cite{10.1145/2484838.2484846} have been proposed to protect personal data.
Differential privacy adds randomness to data, making it hard to identify individuals.
K-anonymity groups individuals with similar characteristics, while L-diversity and t-closeness generalize data to prevent sensitive information disclosure\cite{8329504,10.1145/1217299.1217302, li2006t}.
Secure computation allows for the computation of private data without revealing it. 
While effective at protecting privacy, these methods usually degrade data quality and prediction accuracy.
Therefore, choosing a privacy-preserving method is crucial to balancing privacy and performance.
Additionally, taxi demand data is inherently distributed due to the geographical spread of operations, the presence of multiple independent service providers, and stringent data privacy regulations. Centralizing this data for analysis presents significant challenges, including logistical complexities, resource constraints, and compliance with privacy laws.

Federated learning offers a robust solution by enabling collaborative model training across distributed datasets owned by different entities without necessitating direct data exchange. This method involves training local models on individual datasets and subsequently aggregating the model parameters, rather than the raw data, at a central server. Therefore, in this paper, we adopt federated learning to ensure that sensitive customer information remains secure and under the control of the original data owners while still benefiting from the collective learning process.

\begin{figure}[tb]
    \centering
    \includegraphics[width=8.5cm]{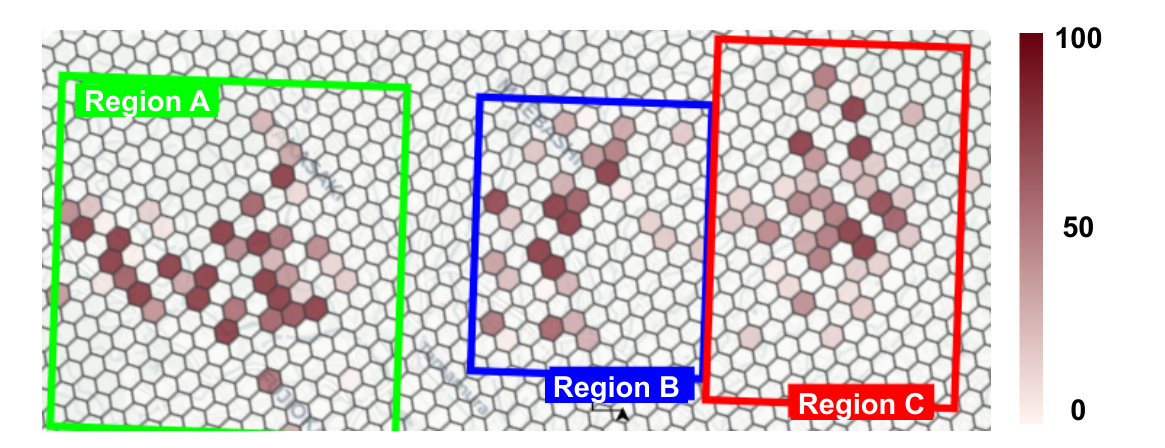}
    \caption{Taxi demand distribution in different regions.}
    \label{fig:imbalance_vis_hex}
\end{figure}

By using federated learning, the system can capture and leverage the unique demand patterns of different regions and providers, leading to more accurate and generalized demand prediction models. However, the application of federated learning in this context faces a generalization problem as the local models are trained with region-specific data. Since taxi demand depends on regional characteristics (e.g., road network) and label distribution also varies, data on taxi demand is heterogeneous among facilities, as shown in Figure \ref{fig:imbalance_vis_hex}. Therefore, each local model in a facility tends to overfit to its region's characteristics, affecting model aggregation in federated learning.

In this paper, we propose \system, a novel system that integrates collaborative learning with contrastive learning to accurately predict taxi demand, even with heterogeneous data among facilities. Our approach ensures high performance by enabling multiple parties to collaboratively train a demand-prediction model through hierarchical federated learning. In this approach, similar parties are clustered together, and federated learning is applied within each cluster to address the non-IID (non-independent and identically distributed) problem. This clustering approach allows the taxi demand prediction model to obtain more comprehensive knowledge about taxi demand and prevents the performance decline that can result from aggregating dissimilar models, leading to more accurate predictions.
We define similarity between different entities based on model similarity rather than data similarity, ensuring both privacy and security. Additionally, contrastive learning is adopted for local training to address class(demand) imbalance and data scarcity problems at each entity.
Furthermore, we employ membership inference attacks in the context of spatial data to validate and ensure that \system\ is robust against such attacks. This validation process confirms that our system not only provides accurate predictions but also maintains strong privacy protections.


We rigorously evaluated the proposed system using real-world data from \textcolor{black}{five} taxi service providers in Japan, collected over fourteen months.
This data assessed the system's ability to maintain prediction performance while preserving passenger privacy. 
The results show that our proposed system achieves \textcolor{black}{2.2}\% higher accuracy than non-federated approaches requiring the sharing of customer data among facilities.



Our contributions are summarized as follows:
\begin{itemize}
    \item We introduce \system, a novel taxi demand prediction system that prioritizes customer privacy by constructing the model without sharing sensitive data. This design achieves privacy protection at the architectural level, ensuring compliance with data privacy regulations.
    \item To the best of our knowledge, this is the first research effort to tackle two critical challenges in taxi demand prediction: (i) data scarcity and imbalance within each region, and (ii) the decentralized, region-specific nature of taxi demand data. Our comprehensive system extracts common features across regions, facilitates inter-regional knowledge sharing, and adapts the model to the unique characteristics of each region, thereby overcoming the limitations of traditional neural networks such as LSTM and conventional federated learning.
    \item We collected a large-scale, real-world dataset from five taxi service providers in Japan over fourteen months and conducted extensive experiments. Our results demonstrate that \system significantly outperforms state-of-the-art methods in terms of prediction accuracy and robustness.
    \item We rigorously validate \system against membership inference attacks to ensure the robustness and security of our approach. Our system prevents overfitting to local data through the integration of collaborative and contrastive learning, thereby maintaining high prediction performance while safeguarding data privacy.
\end{itemize}

This paper is organized as follows:
Section \ref{sec:system} presents our methodology, including the historical taxi demand encoder, the architecture of contrastive learning, and the decentralized collaborative learning mechanism.
Section \ref{sec:evaluation} discusses the experimental setup and analyzes the results.
Section \ref{sec:related_work} reviews related work in taxi demand prediction, privacy protection methods for mobility data, countermeasure of class imbalance, and federated learning.
Finally, Section \ref{sec:conclusion} concludes the paper.

\begin{figure}[tb]
    \centering
    \includegraphics[width=8.5cm]{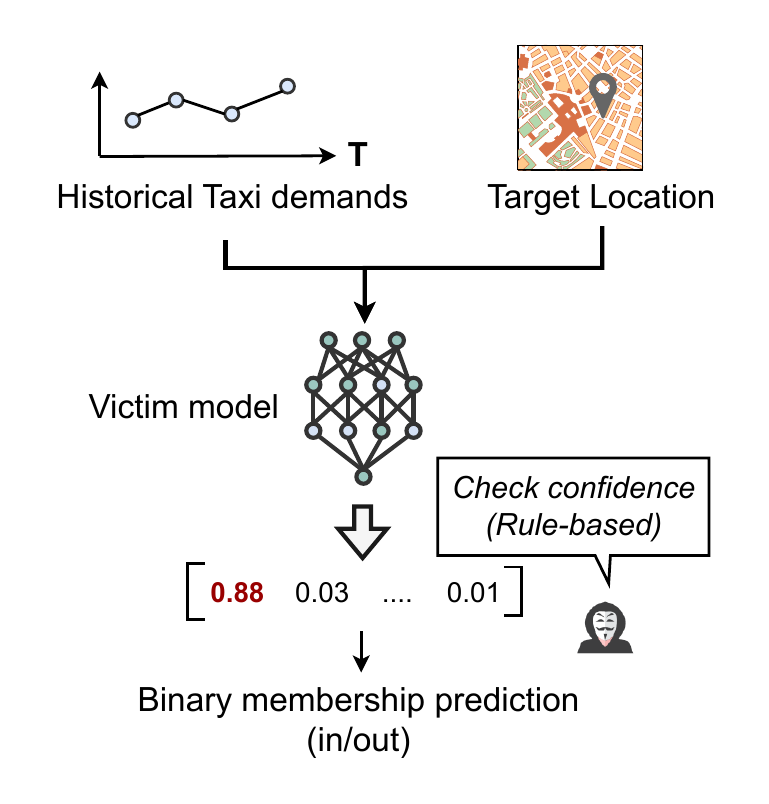}
    \caption{The procedure of membership inference attack for taxi demand prediction.}
    \label{fig:Distributed_taxi_demand_prediction_MIA}
\end{figure}

\section{Threat Model}
\label{sec:threat_mia}
Our proposed system uses a machine learning model to predict taxi demand accurately. 
However, machine learning models generally memorize training data details and can leak privacy-sensitive information, posing serious risks.
One of the most severe risks is membership inference attacks, determining whether specific data points were part of the training set\cite{shokri2017membership,zhang2020locmia}.
This is particularly concerning with sensitive information, such as personal or spatio-temporal data, as it could lead to identifying and tracking individuals.

A membership inference attack (MIA) aims to determine whether a specific individual's data was used to train a machine learning model.
This attack poses a significant threat to the security and privacy of machine learning models\cite{shokri2017membership,zhang2020locmia}.
Shokri et al. \cite{shokri2017membership} specifically evaluated the efficacy of MIA in computer vision, particularly in scenarios where the attacker lacks prior knowledge of the data used to train the model.

This work extends the investigation of MIAs to the context of taxi-demand prediction models, which are trained using mobility data.
In particular, these models are trained using aggregated trajectory data, and an attacker could exploit the MIA to ascertain whether a specific individual's data (i.e., their taxi ride data) was used to train the model even from aggregated data as demonstrated in \cite{zhang2020locmia}.
Consequently, a successful MIA could enable the attacker to infer sensitive information about individuals, such as travel habits or daily routines.
For example, knowing that an individual's data was used for training could allow the attacker to identify them as a frequent user of taxis to specific locations at particular times. 
This information could be used to track the individual's movements and infer private details such as work schedules, residential locations, and health-related activities.

MIA can be implemented in a white-box setting, which allows the attacker to access the model's architecture, parameters, and training data.
This is a strong assumption and cannot be generally realized in practice.
We employ a black-box version of MIA to ensure practical use cases where the attacker cannot access the model's architecture, parameters, and training data to evaluate the robustness of our proposed system.
The procedure of the MIA attack is illustrated in Fig\ref{fig:Distributed_taxi_demand_prediction_MIA}.
The attacker initially collects a set of unlabeled data of the input (location, time, historical taxi demand distribution) but not the output (demand)).
This data would then be fed to the victim prediction demand model, which would be queried repeatedly to analyze the model's predictions.
In this work, we assume the taxi demand prediction model aims to classify taxi demand (e.g., low, mid, and high).
The attacker infers whether a specific individual's data is used to train the model, leveraging the confidence of the target model.
Since the victim demand prediction model outputs the probability of the demand class and the output probabilities are strongly biased if the input data are included in the training dataset, an attacker can detect "member" or "non-member" by leveraging the probabilities of the target model.

\section{Proposed system}
\label{sec:system}

\begin{figure*}[!tb]
    \centering
    \includegraphics[width=16cm]{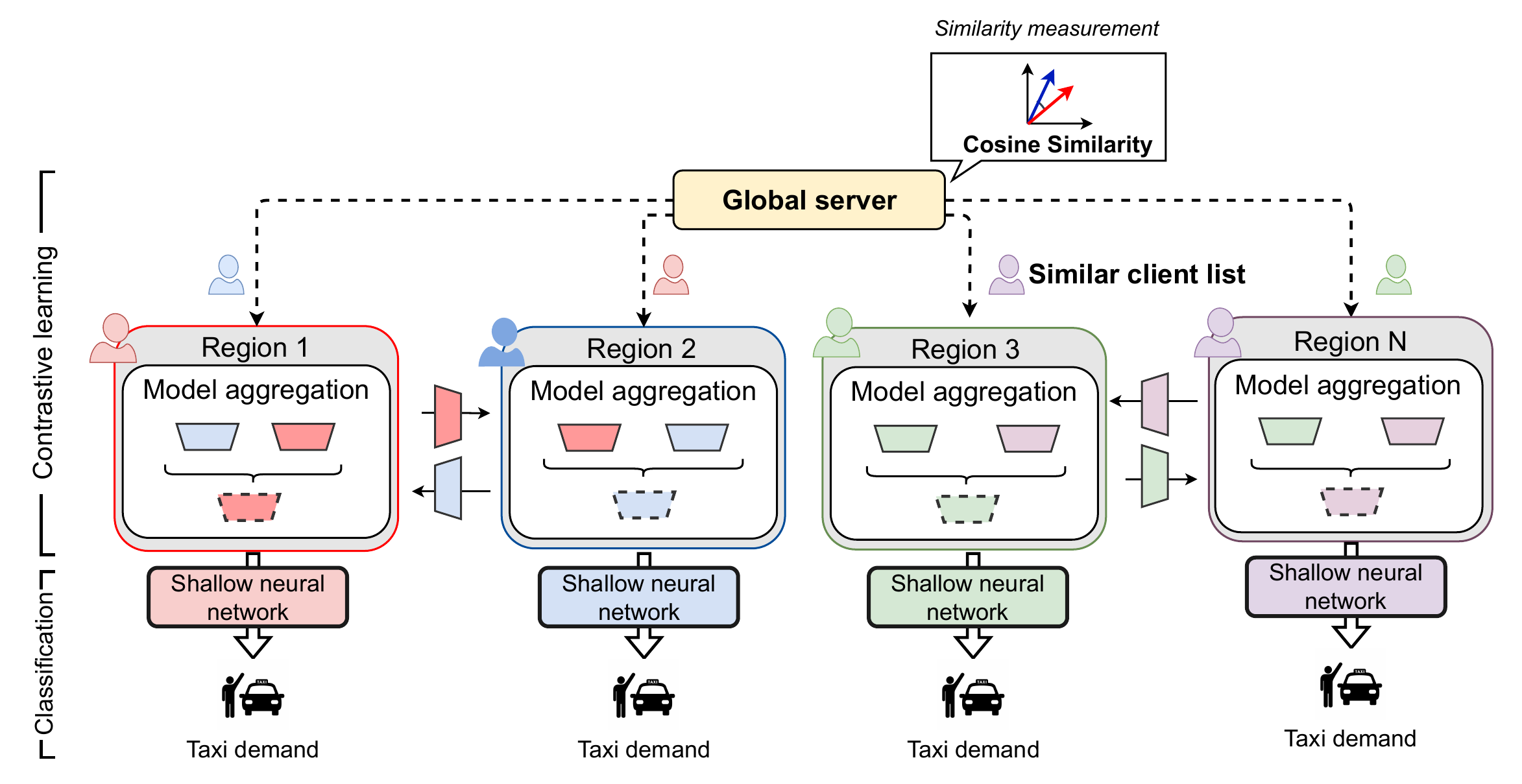}
    \caption{\system overview in heterogenous environments}
    \label{fig:taxi demand_crest_system_architecture}
\end{figure*}

The objective of \system is to predict taxi demand using the model of each client in their respective regions.
To preprocess raw data such as historical taxi demand into a manageable and interpretable format for machine learning models, the \textbf{hexagonal virtual gridding module} converts all data into hexagonal grid cell formats.
The formatted data is then input into the \textbf{feature extractor module} of each client to extract meaningful latent features from complex spatio-temporal input within their respective regions.
This extraction is achieved using contrastive learning techniques, as described in \cite{chen2020simple}, where training is conducted self-supervised.
Contrastive learning is designed to reduce the distance between the representations of different augmented data of the same input (i.e., positive pairs) and increase the distance between the representations of augmented views of different input (i.e., negative pairs).
Therefore, contrastive learning contributes to distinguishable latent representation, even self-supervised.
Then, clients communicate to efficiently share their knowledge, including the weights of the contrastive learning models, through a \textbf{decentralized collaborative learning mechanism}.
Within this mechanism, clients can identify and collaborate with clients with similar data distributions, enabling efficient collaborative learning.
This mechanism identifies similar clients by measuring the similarity of the local model's update because the similar direction of model update helps convergence of aggregated model and reflects similar data distribution as written in \cite{9954190}.
Finally, each client fine-tunes a shallow neural network, the \textbf{client-tailored classifier}, to predict taxi demand within their respective regions.
This fine-tuning process utilizes the feature extractor trained in the previous module.
It is noteworthy that the virtual gridding module, feature extractor module, and client-tailored classifier are client-dependent functions that run on each client, while the decentralized federated learning mechanism is run on a single global server.
The architecture will be explained later in this section.

Since our focus is on a non-IID and class-imbalanced taxi demand dataset, 
\system needs to obtain comprehensive knowledge about taxi demand while personalizing the prediction model to predict taxi demand accurately in the target region.
\system utilizes contrastive learning in the feature extraction phase to derive general knowledge about taxi demand, as it does not rely on label information.
This enables us to obtain meaningful representation without feeding each component's class imbalance ratio, which sets it apart from existing methods for class-imbalanced federated learning\cite{wang2020addressing}.
Furthermore, \system can enhance the efficiency of the learning process for each client and personalize each model based on the similarity assessment of clients and corresponding grouping.

\color{black}

\subsection{Hexagonal Virtual Gridding}\label{sec:hex_grid}
This Module is a pivotal component of the data generation system aimed at accurately computing the taxi demand for inputs of machine learning models. 
This module plays a vital role in transforming raw trajectory data into a manageable and interpretable format suitable for machine learning models' utilization.
The primary objective of this module is to partition the map into evenly spaced hexagonal cells, wherein each cell represents a distinct area on the map.
Subsequently, the module calculates the number of demand events that transpired within each hexagonal cell during a specific time-slot in a day.
Notably, the module does not differentiate between pick-up and drop-off events, solely focusing on the aggregate count of demand events within each cell.
The gridding process entails superimposing a virtual hexagonal grid onto the map.
This approach enables the system to provide a comprehensive overview of taxi demand in different city areas, facilitating predictions regarding the number of demand events in distinct areas.
Moreover, it allows for the facile visualization of demand patterns and identifies regions characterized by high or low demand.

We adopt a hexagonal grid instead of a square grid due to its efficiency and effectiveness in representing geographic regions compared to squares.
Specifically, hexagons provide balanced neighboring as each shares a common edge with six neighboring hexagons.
This property ensures a more equitable distribution of neighboring cells, reducing edge effects and accurately representing spatial relationships.
Additionally, hexagons allow for more compact packing, covering a given area with fewer cells, resulting in a more accurate depiction of the geographic space and reduced redundancy within the grid.
Moreover, hexagonal cells have equidistant centers, ensuring consistent and regular spacing throughout the grid.
This characteristic facilitates precise distance calculations and enables robust spatial analysis. 
Furthermore, hexagons offer directional flexibility, allowing movement in six possible directions. 
This flexibility enhances the system's ability to capture and analyze spatial patterns, making it particularly advantageous in modeling transportation systems and understanding travel patterns.

\subsection{Feature Extractor Module}
\subsubsection{Feature encoder}
Taxi demand depends on Spatio-temporal characteristics.
To capture the spatio-temporal feature, we employ a feature encoder that receives the time series data (\textit{i.e.}, historical taxi demand) and the characteristics of a target region.
This feature encoder captures past demand trends and spatial distributions, providing valuable insights for predicting future demand. 
As shown in Figure \ref{fig:taxi demand_crest-feature_encoder}, we extract the feature from a sequence of historical taxi demand using Transformer \cite{liu2021convtransformer} and predict future taxi demand.

\begin{figure}[tb]
    \centering
    \includegraphics[width=8.5cm]{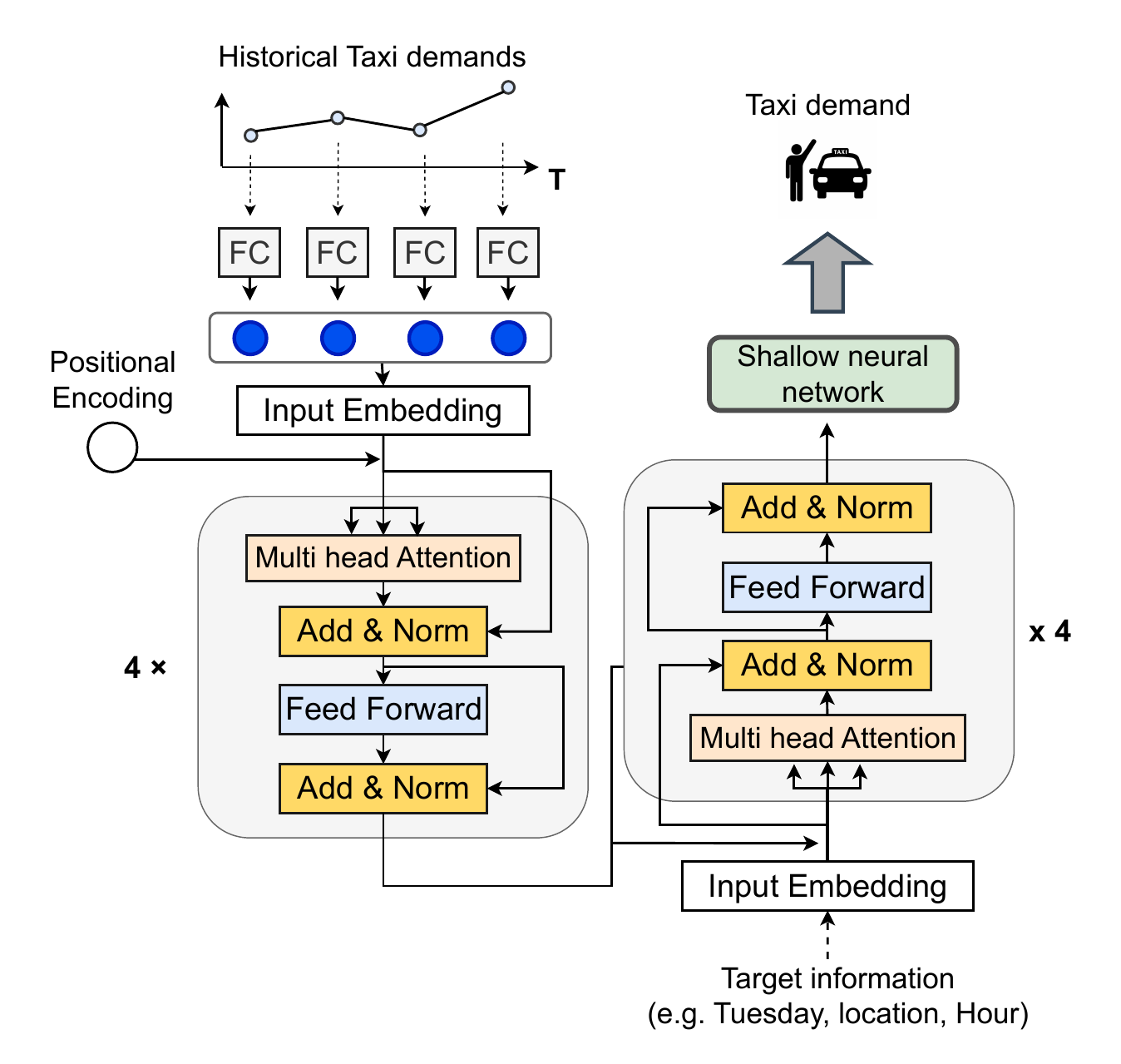}
    \caption{Taxi demand prediction model in our proposed system}
    \label{fig:taxi_demand_prediction_model}
\end{figure}

\color{black}

\subsubsection{Feature Extractor with Contrastive Learning}
\begin{figure}[t]
    \centering
    \includegraphics[width=8.5cm]{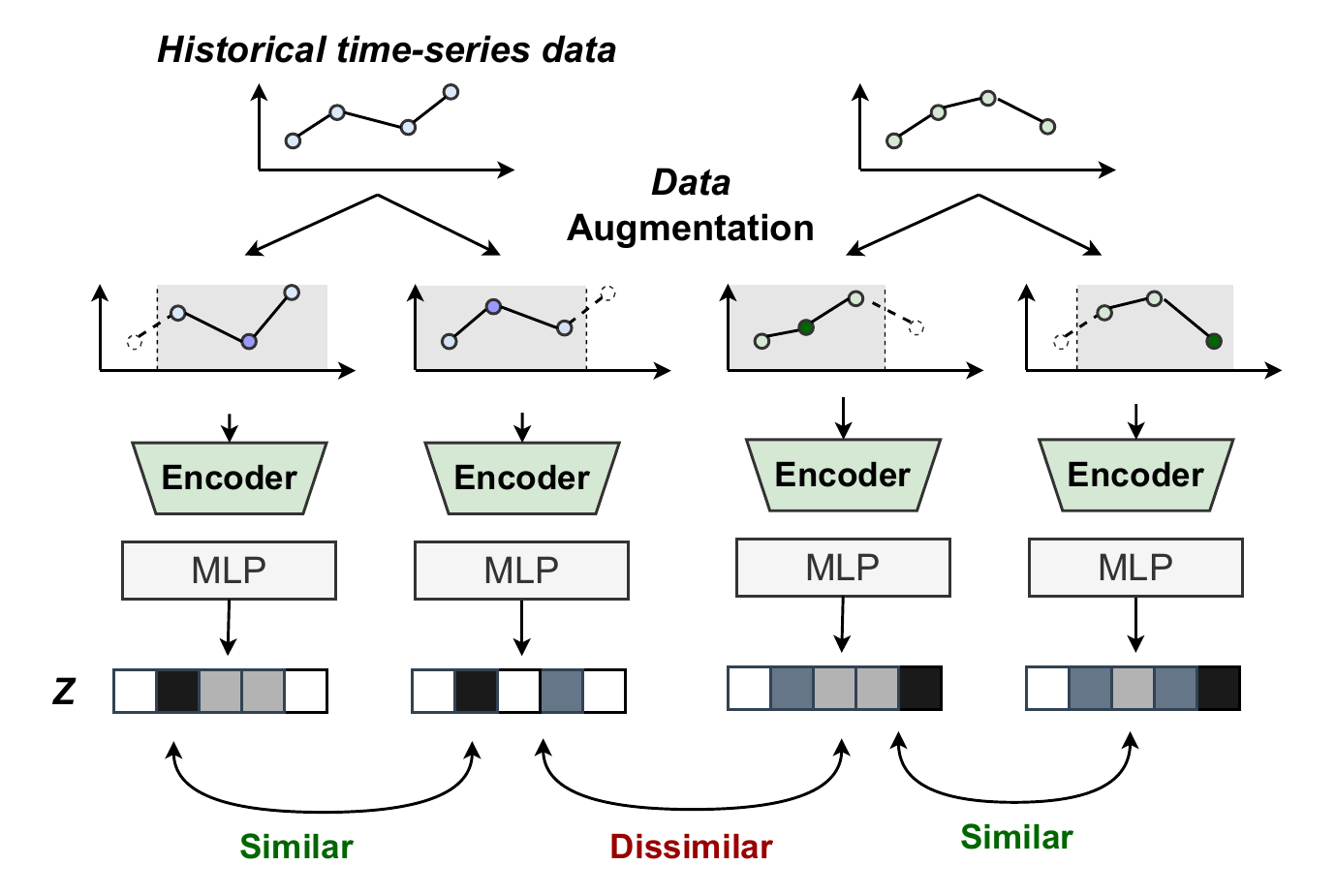}
    \caption{Contrastive learning illustration. The input data are twice augmented by both cropping and adding noise. The generated pair from the same data is encoded into a similar representation, but the representations from different data are kept away in latent space.}
    \label{fig:taxi demand_crest-contrastive_learning}
\end{figure}

This part obtains meaningful representation from the spatio-temporal inputs in grid-structured data.
To avoid the effects of class imbalance and get feasible representation from complex spatio-temporal features, we incorporate a contrastive learning mechanism. 
Inspired by recent contrastive learning such as SimCLR\cite{chen2020simple}
in the computer vision research domain, our feature extractor learns representations by maximizing agreement between differently augmented views of the same data example via a contrastive loss in the latent space.
As shown in Figure \ref{fig:taxi demand_crest-contrastive_learning}, this framework comprises four major components.

\begin{itemize}
    \item Stochastic data augmentation: From each data sample, it randomly generates two correlated views of the sample, denoted as  $\Tilde{x}_{\alpha}$ and $\Tilde{x}_{\beta}$.
    They are considered a positive pair.
    In \system, we apply two types of augmentations: random cropping in terms of time series and noising.
    \item Encoder $f(\cdot)$: It extracts representation vectors from the data augmented in the previous component.
    We employ an encoder that consists of two types of view to obtain $h_{\alpha} = f(\Tilde{x}_{\alpha})$ where $h_{\alpha} \in \mathbb{R}^{d}$ is a $d$-dimensional output of the encoder.
    
    \item Small neural network projection head $g(\cdot)$: It maps representations to the space where contrastive loss is applied.
    We use a shallow neural network (MLP) with one hidden layer to obtain $z_{\alpha} = g(h_{\alpha})$.

    \item Contrastive loss prediction: A contrastive loss function is defined for this contrastive prediction task, given a set $\{\Tilde{x}_k\}$ including a positive pair ${\Tilde{x}_{\alpha}}$ and ${\Tilde{x}_{\beta}}$ of samples.
\end{itemize}

We randomly pick up $B$ examples for a minibatch and generate $2B$ augmented data using stochastic data augmentation. 
Then, we defined the contrastive prediction task on the augmented data, where for each positive pair, we apply contrastive learning regarding the rest $2(B-1)$ augmented examples as negative examples. 
Let $\text{sim}(u, v) = u^{\mathsf{T}}v/||u||\cdot||v||$ denote the dot product between two normalized $u$ and $v$ (i.e. cosine similarity).
Then the loss function for a positive pair of examples $(\Tilde{x}_\alpha, \Tilde{x}_\beta)$ is defined as:

\begin{equation}
    \ell_{\alpha, \beta} = -\log \frac{\exp(\text{sim}(z_{\alpha}, z_{\beta})/\tau)}{\sum_{k \in 2B\_index\_set} b_{[k \neq \alpha]} \exp(\text{sim}(z_{k}, z_{\beta})/\tau)}
\end{equation}
where the value of $b_{[k \neq \alpha]} \in \{0, 1\}$ is 1 if and only if  $k \neq \alpha$, and $\tau$ denotes a temperature parameter.
The final loss is computed across all positive pairs, both $(\Tilde{x}_{\alpha}, \Tilde{x}_{\beta})$
and $(\Tilde{x}_{\beta}, \Tilde{x}_{\alpha})$ in a mini-batch. 
This loss has been used in \cite{chen2020simple} and called \textit{NT-Xent loss (the normalized temperature-scaled cross-entropy loss)}.

\subsection{Decentralized Collaborate Learning Mechanism}
\label{sec:knowledge_sharing_mechanism}

\begin{figure}[tb]
    \centering
    \includegraphics[width=8cm]{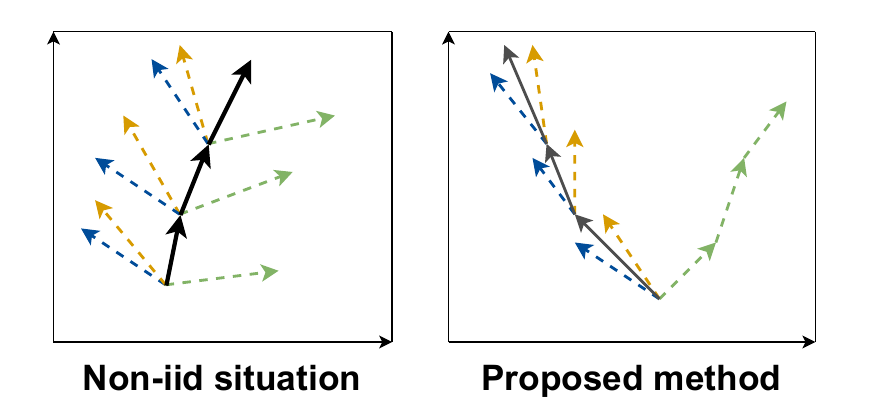}
    \caption{Visualization of optimization path for normal FL(a) and Proposed method(b) in non-iid situation.}
    \label{fig:model_weight_update_direction}
\end{figure}

This section presents a decentralized, collaborative learning approach that ensures each client's performance.
Aggregating models trained on different data distributions can typically degrade the performance of a global model.
To maintain high performance in these non-IID settings, \system incorporates a similar neighbor selection mechanism \cite{9954190}.
This mechanism identifies clients with similar data distributions, assisting the performance of each model as illustrated in Figure\ref{fig:model_weight_update_direction}.
The personalized knowledge-sharing mechanism in \system involves two stages: \textit{similar client selection} and \textit{distributed model update}.

\subsubsection{Similar Client Selection}
In this stage, \system aims to identify similar clients for each local client.
Each local client is denoted as $i$ and possesses a model with model parameters $w_i$.
These parameters are updated over $E$ local epochs and uploaded to a \textit{global server} for $T_{init}$ rounds.
This procedure mirrors the commonly used approach in centralized federated learning.

We measure the similarity of clients using the Personalized Adaptive Neighbor Matching (PANMGrad) \cite{9954190} method, which does not require setting the number of clusters.
The global server calculates the similarity of models of two clients $i$ and $j$ using a combination of two parameters.
The first parameter is given in Eq. (\ref{eq:model_similarity_g}), where $g^{t}_{i}$ is the vectorized gradient of client $i$ in round $t$ and is expected to represent the data distribution in client $i$. $g^{t}_{i}$ is obtained by $g^{t}_{i} = w^{t}_{i} - w^{t-1}_{i}$, where $t$ is the current time round.
We note that $w^{t}_{i}$ is initialized by the same global model at the beginning of local training in each round, and one-round update $g^{t}_{i}$ can be noisy.

To capture the historical optimization directions of each model, we also introduce accumulated weight updates from the initial model, $h^{t}_{i} = w^{t}_{i} - w^{0}_{i}$, and use it in the second parameter given by Eq. (\ref{eq:model_similarity_h}). $h^{t}_{i}$ is the accumulated vectorized gradient of client $i$ in round $t$.
The global server calculates the similarity of models of two clients using similarity metrics given in Eq. (\ref{eq:model_similarity}), where $\gamma \in [0,1]$ is a hyperparameter.

\begin{align}
\label{eq:model_similarity_g}
    G_{i,j} &= \frac{<g^{t}_{i},g^{t}_{j}>}{||g^{t}_{i}||\cdot ||g^{t}_{j}||} \\
\label{eq:model_similarity_h}
    H_{i,j} &= \frac{<h^{t}_{i},h^{t}_{j}>}{||h^{t}_{i}||\cdot ||h^{t}_{j}||} \\
\label{eq:model_similarity}
    \text{sim}_{i,j} &= \gamma G_{i,j} + (1-\gamma) H_{i,j} 
\end{align}

The global server selects the clients whose similarity is higher than the threshold $m$ as the ``neighbor clients'' using average similarity during $T_{init}$.
The threshold $m$ is the system parameter.
Our client section method is summarized as Algorithm\ref{alg1} of Figure \ref{fig:scs}, where $N$ is the number of clients.

\begin{figure}[t]
\begin{algorithm}[H]
    \caption{Similar clients selection for clients}
    \label{alg1}
    \begin{algorithmic}[H]
    \Function{main}{}
    \State $\textbf{w\_record} \leftarrow \{\}$
    \For{$t \hspace{0.2cm} \text{in} \hspace{0.2cm} T_{init}$}
        \For{$i \hspace{0.2cm} \text{in} \hspace{0.2cm} N$}
            \State $w^{t}_i \leftarrow \text{client\_update}(i)$
            \State $\textbf{w\_record}.add(w^{t}_i)$ 
        \EndFor
    \EndFor
    \State $R \leftarrow \text{calculate similarity using \textbf{w\_record}}$

    \For{$round \hspace{0.2cm} \text{in} \hspace{0.2cm} T$}
        \For{$i \hspace{0.2cm} \text{in} \hspace{0.2cm} N$}
            \State $\textbf{similar\_clients}_i \leftarrow \text{get\_similar\_clients}(i, R)$
            \State $\text{train \& communicate with}$ $\textbf{similar\_clients}_i$
        \EndFor
    \EndFor
    \EndFunction
\end{algorithmic}
\end{algorithm}
\caption{Similarity-based Client Selection Algorithm.}
\label{fig:scs}
\end{figure}

\subsubsection{Distributed Model Update}
In this stage, each client communicates with similar clients in the previous stage.
More concretely, the model parameters $w_i$ of client $i$ are delivered to each client $j$, which is a similar client of $i$.
Client $i$ then waits for the models sent from up to $Nsim_i$ clients and aggregates these received models.
We use $n$ to represent the number of received models, and it is important to note that $n \leq N_{\text{sim}_i}$, taking into account cases where model delivery may fail or not be performed for some reasons.
Eq. (\ref{eq:aggregation}) is the federated aggregation of the received models.

\begin{equation}
\label{eq:aggregation}
    \bar{w}^{t+1}_{i} \leftarrow \frac{1}{n+1}\sum_{j \in \{\textrm{similar clients of } i\} \cup \{i\} }w_j^{t} 
\end{equation}

The new aggregated model $\bar{w}_{i}^{t+1}$ is then trained for $E$ epochs before it is ready to be gossiped again.

\subsection{Client Tailored Classification}
This module takes the output (\textit{i.e.}, latent features) of the feature extractor module as input and classifies the given inputs into some classes.
This work aims to classify taxi demand into three classes: high, mid, and low.
It is two layers MLP and trained by a cost-sensitive loss function of Eq. (\ref{eq:weighted_class_loss}).

\begin{equation}
\label{eq:weighted_class_loss}
    \ell(p, q) = -\sum_{x=1}^{N} \frac{1}{w_x}p(x)\log(q(x))
\end{equation}

The loss function calculates the loss according to the ratio of the classes $w_x$, where $p$ and $q$ are the actual label and output probability of the model, respectively.

\color{black}



\section{Evaluation}
\label{sec:evaluation}
\subsection{Data collection and setup}
\label{sec:data_collection}

\subsubsection{Data collection}

We gathered real-world data from \textcolor{black}{five} service facilities in Japan over fourteen months.
The collected data includes (1) vehicle information and their trajectories (including idle time) and (2) spatiotemporal data of each customer’s pickup and drop-off event for each vehicle.
The system determined the trajectory of each customer’s trip by merging the two datasets using the vehicle ID and time as the key factors.
This resulted in \textcolor{black}{around 17000} trips, with taxi demands ranging from 0 to \textcolor{black}{20}, calculated using a grid size of around 1 km and a time slot of 1 hour.
The trajectory data was obtained through GPS for latitude and longitude, with data acquisition intervals of approximately every 5 seconds, with some missing data.
To determine the locations of pickup and drop-off events, we used data on vehicle positions during the 45 seconds before and after the event, if available.
The event was omitted from the evaluation data if the data was absent.
Table\ref{tab:ourdataset} summarizes the information for our datasets.

\begin{table}[tb]
    \centering
    \caption{Our Dataset description.}
    \label{tab:ourdataset}
    \begin{tabular}{c||ccp{1.2cm}} \hline
        Dataset & Duration & Area[km $\times$ km] & \#records \\ \hline
        \textbf{r1} &  \quad (5 months) & $36.6 \times 38.6$ & 6.5k \\ 
        \textbf{r2} &  \quad (14 months) & $16.6 \times 13.5$ & 2.6k \\ 
        \textbf{r3} &  \quad (5 months) & $14.4 \times 14.4$ & 6.7k \\
        \textbf{r4} &  \quad (5 months) & $31.1 \times 31.5$ & 1.6k \\ 
        \textbf{r5} & \quad (5 months) & $16.6 \times 11.7$ & 2.9k \\ \hline
    \end{tabular}
\end{table}

\color{black}

\begin{table}[!t]
    \centering
    \caption{Experimental Settings.}
    \begin{tabular}{l|l}
    \hline
       Parameter & \begin{tabular}{c} Explored range\\(\textbf{Bold}: default value)\end{tabular} \\ \hline
        Cell edge length [km] & \{$\bm{1}$, 2\} \\
        Time interval [hour] & \{0.5, $\bm{1}$, 2\} \\
        Batch size & \{8, 16 , 32, 64, $\bm{128}$\} \\
        Temperature $\tau$ & \{0.3, $\bm{0.5}$, 0.8\}\\
        Initial communication round $T_{init}$ & \{1, 2, 5, $\bm{10}$\} \\
        Communication round $T$& \{10, 20, $\bm{30}$\} \\
        Local epoch $E$& \{$\bm{1}$, 3, 5\}\\
        Similarity threshold $m$ & \{-1, $\bm{-0.5}$, 0, 0.5, 0.8\} \\
        Parameter of similarity metrics $\gamma$ & \{0.2, $\bm{0.4}$, 0.6, 0.8\} \\
        Model size of \system & \{0.76 Million\} \\
        \hline
    \end{tabular}
    \label{tab:config}
\end{table}

\color{red}

\color{black}
\subsubsection{Metrics}
We focus on balanced accuracy for evaluating taxi-demand predictions because the gathered data are class-imbalanced. Conventional accuracy metrics are insufficient for this data type, as they can be misleading when classes are not equally represented. Therefore, we adopt balanced accuracy, which provides a more reliable measure by averaging the accuracy across all classes, ensuring that each class is equally considered in the evaluation.

\begin{figure}[tb]
    \centering
    \includegraphics[width=8.5cm]{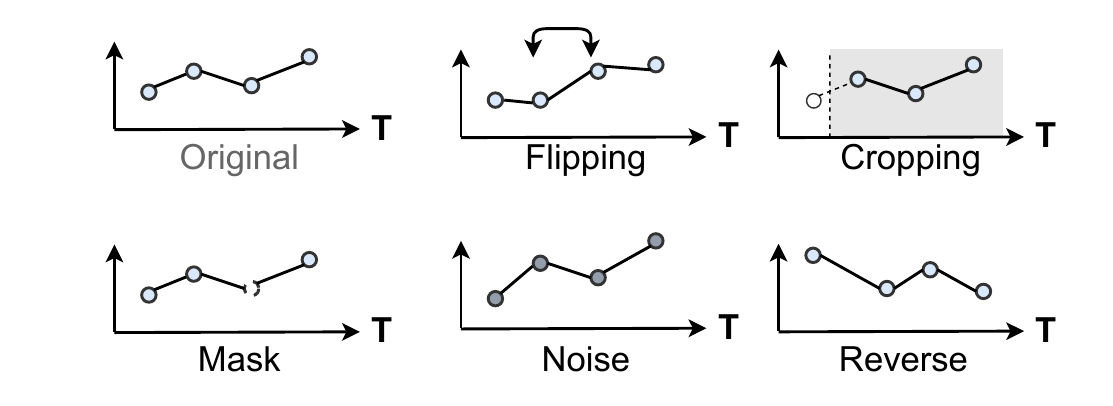}
    \caption{Data augmentation techniques.}
    \label{fig:timeseries_augmentaion_explanation}
\end{figure}

\subsection{System parameter analysis}

\color{black}
\subsubsection{Data augmentation technique}

To obtain comprehensive and general knowledge about taxi demand prediction, \system\ incorporates time-series data augmentation techniques in the contrastive learning phase.
Since the diversity and quality of augmented data significantly influenced the performance of the feature extract encoder, the data augmentation technique is crucial for \system.
Thus, we consider several common augmentations to investigate the impact of stochastic time-series data augmentation techniques systematically.
To understand the effects of individual data augmentations and the importance of augmentation composition, we investigate the performance of our framework when applying various augmentations: flipping, noise, masking, and cropping.

Adding noise to data is widely used for time-series data\cite{wickstrom2022mixing, yeh2023toward}, which adds noise to each data.
In Masking, we masked data of randomly selected points, which is 10\% of time-series data in this work.
Cropping is a method of cutting out only a subset of the original data, which can keep the order of each data point and sequential information.
Flipping is swapping some points within sequential data.

Adding noise to data is a common technique for time-series data, which involves introducing random variations to each data point\cite{wickstrom2022mixing, yeh2023toward}.
``Masking'' randomly selects and removes 10\% of the time-series data points in this work.
``Cropping'' extracts a subset of the original data while preserving the order and sequential information of the data points.
``Flipping'' swaps certain points within the sequential data.
``Reverse'' reverses the order of sequential data.
Our data augmentation techniques are summarized in Figure\ref{fig:timeseries_augmentaion_explanation}.

\begin{figure}[tb]
    \centering
    \includegraphics[width=8.5cm]{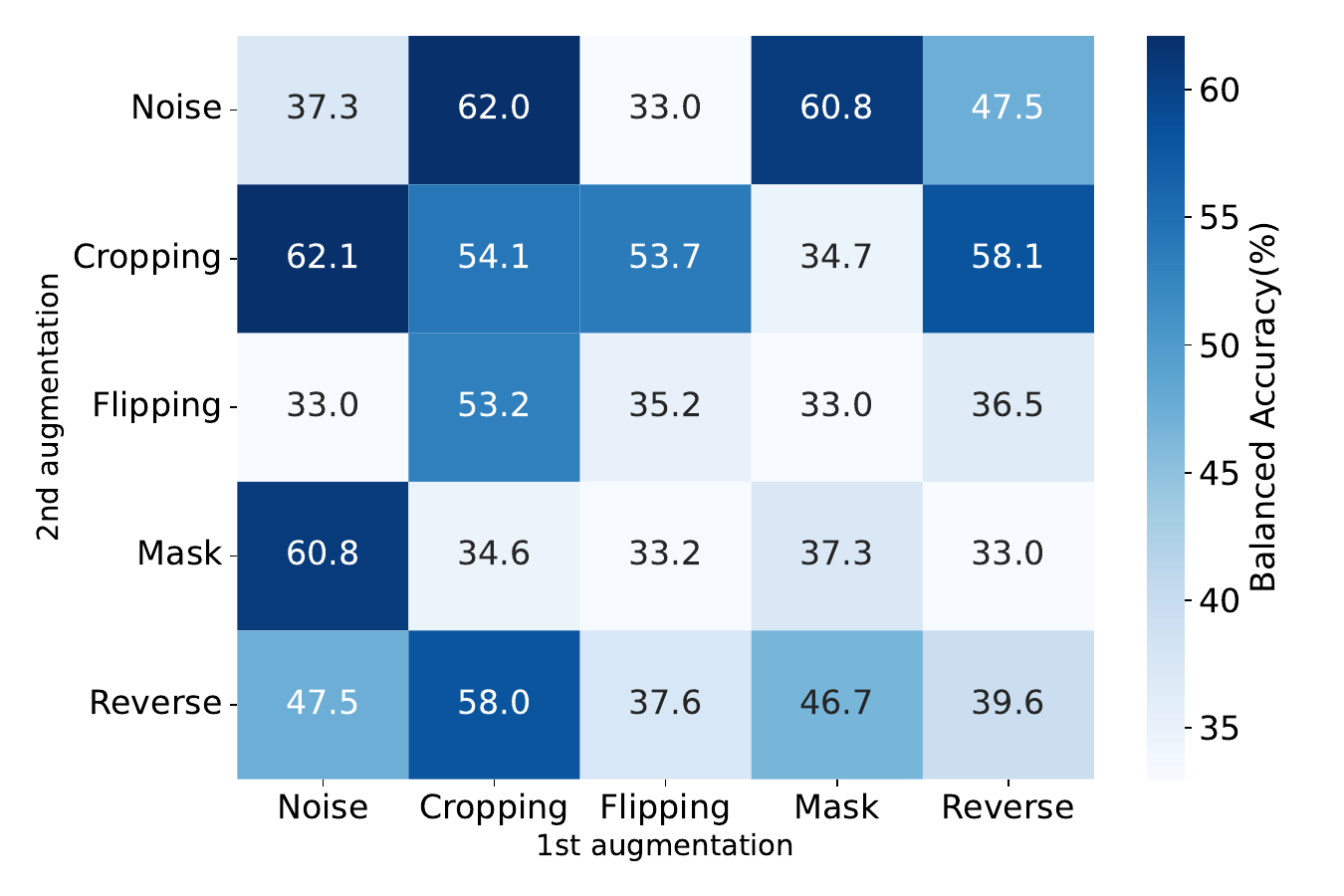}
    \caption{The impact of stochastic data augmentation technique. The number in this figure represents the balanced accuracy of taxi demand prediction.}
    \label{fig:aug_result}
\end{figure}

As demonstrated in \cite{chen2020simple}, multiple stochastic data augmentation is effective for contrastive learning to obtain comprehensive knowledge.
Therefore, we applied two data augmentation at once and evaluated the effectiveness of each combination through the experiments.
Figure\ref{fig:aug_result} displays the taxi demand prediction accuracy for each data augmentation combination.

Figure\ref{fig:aug_result} shows that ``noise\&cropping'' achieves the highest accuracy with these augmentation methods, which indicates that ``reverse'' and ``flipping'' compromise the information for time-series feature and this sequential information such as the order of each point is crucial for taxi demand prediction.
Therefore, in the following evaluation sections, we evaluate our proposed method using ``noise\&cropping'' as a data augmentation technique.

\subsubsection{Similarity thereshold}
\begin{figure}[tb]
    \centering
    \includegraphics[width=8cm]{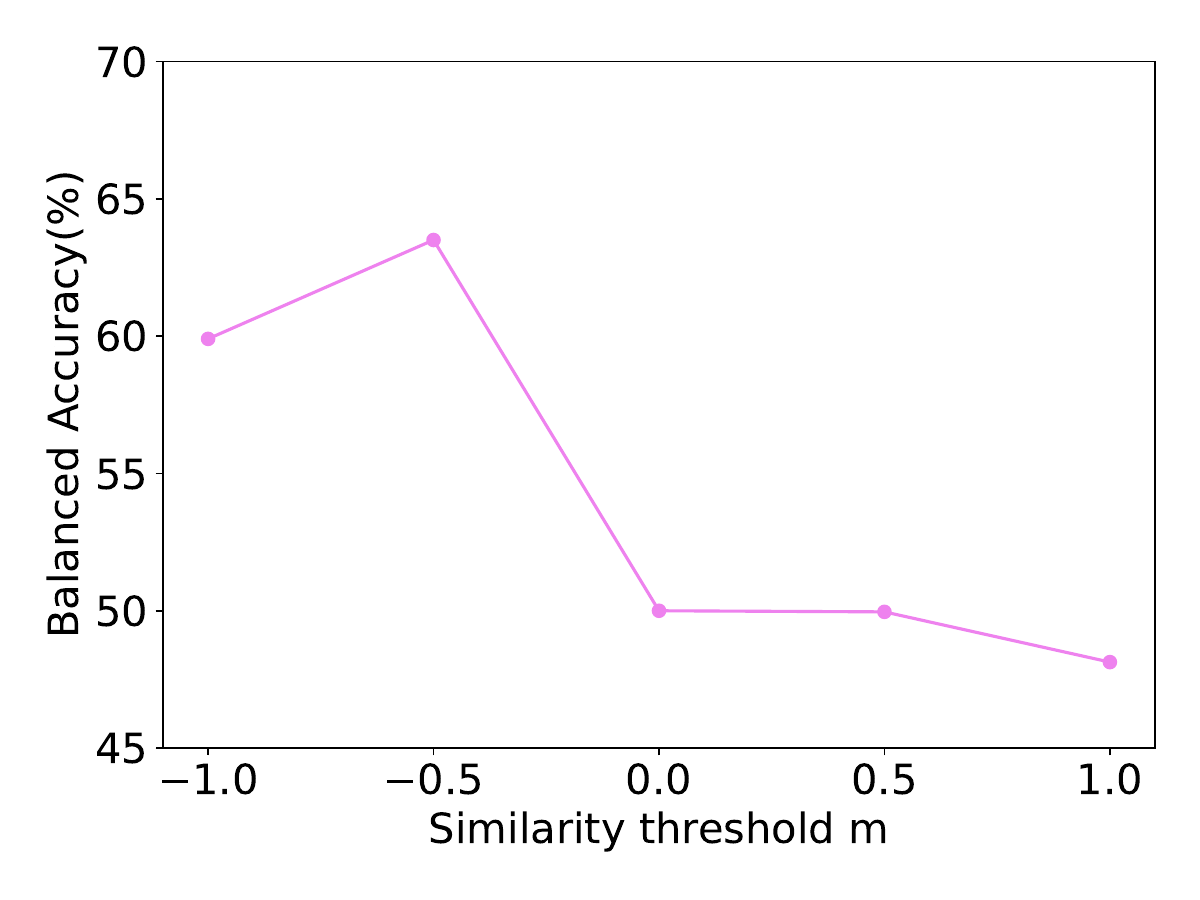}
    \caption{The impact of similarity thresholds.}
    \label{fig:performance_hyper-parameter}
\end{figure}

In \system, for each client, similar clients are selected to share the trained models, and this selection is controlled by $m$, the similarity threshold parameter. 
A smaller value of $m$ groups dissimilar clients and increases the impact of non-IID data distribution.
Meanwhile, with a large value of $m$, the precision may also decrease because clients cannot collect models from sufficient clients to obtain comprehensive knowledge about landslides in decentralized, federated learning.
Therefore, we study the effect of $m$ and consider the appropriate value through the experiment.
Figure \ref{fig:performance_hyper-parameter} illustrates the relationship between the similarity threshold and the balanced accuracy of taxi demand prediction. 
We can observe that \system performs best with $m=-0.5$, and the performance of \system is at the same level as the local model mentioned in the section.\ref{sec:eval_comparison_fl} because each local model almost does not communicate with other models.

\begin{figure*}[tb]
  \begin{minipage}[b]{0.31\linewidth}
    \centering
    \includegraphics[width=5.5cm]{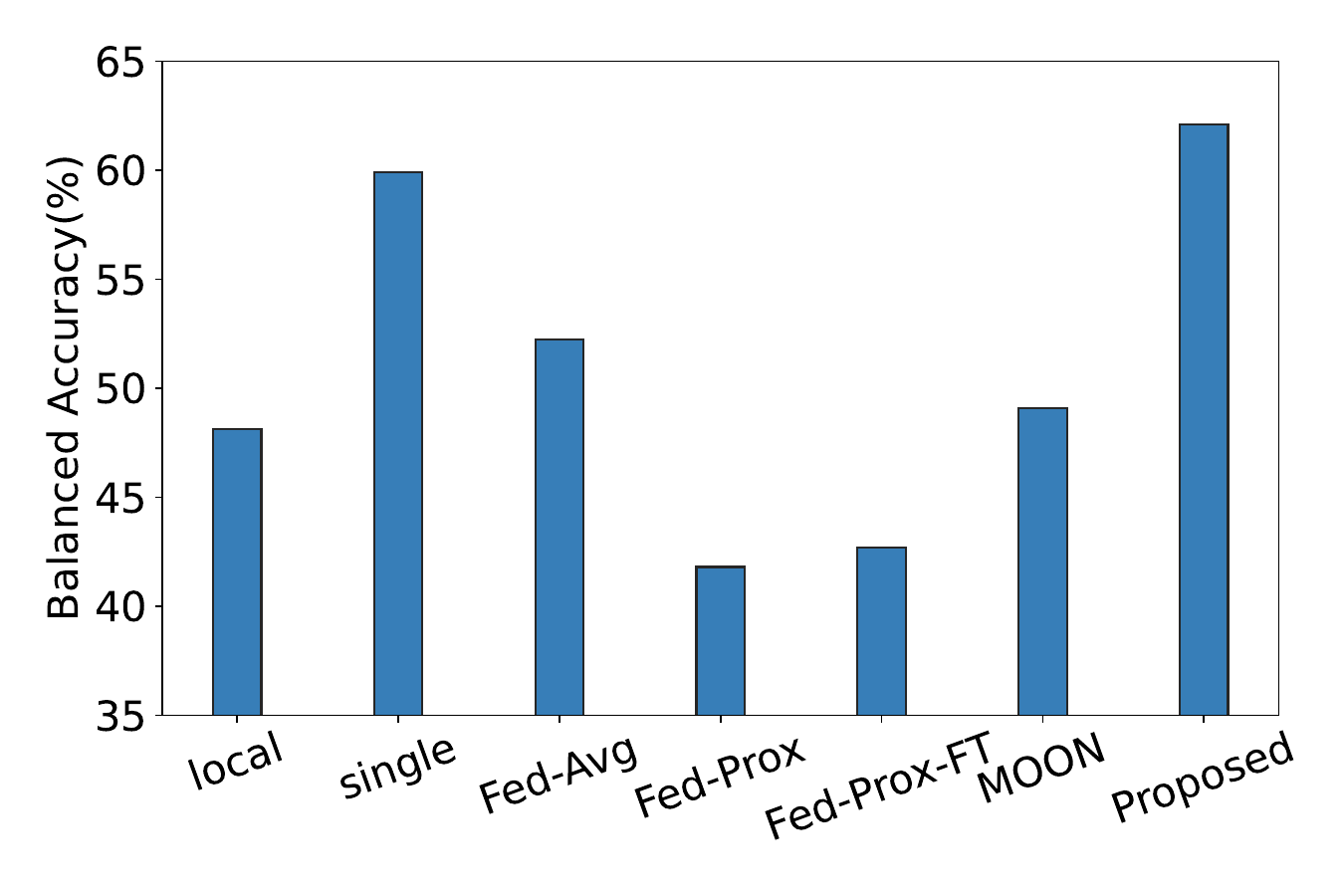}
    \caption{Performance comparison with other distributed learning methods}
    \label{fig:performance_comparison}
  \end{minipage}
  \hspace{0.04\columnwidth}
  \begin{minipage}[b]{0.31\linewidth}
    \centering
    \includegraphics[width=5.5cm]{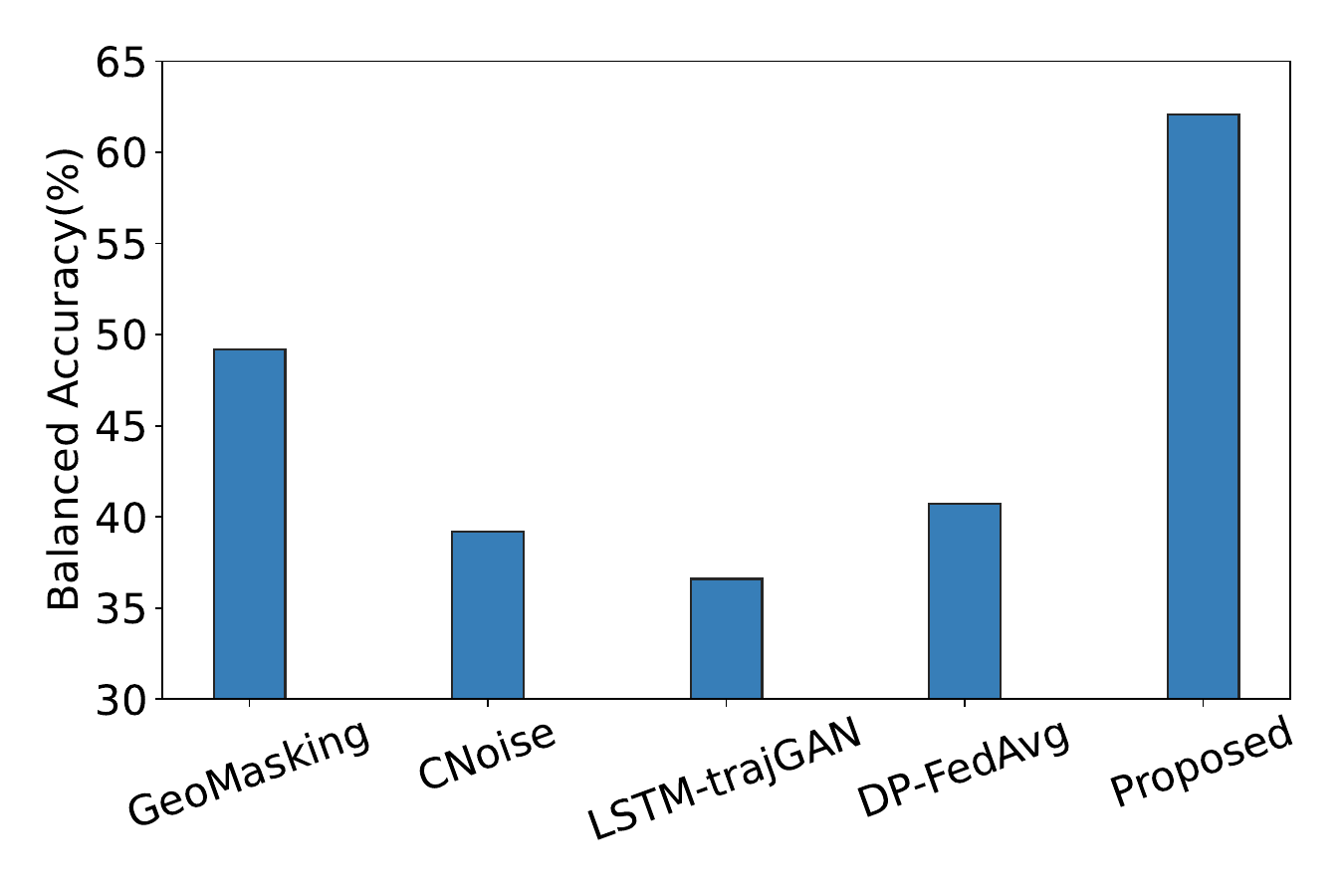}
    \caption{Performance comparison with other privacy preserved methods for data.}
    \label{fig:Comparison_architecture_result}
  \end{minipage}
  \hspace{0.04\columnwidth}
   \begin{minipage}[b]{0.31\linewidth}
    \centering
    \includegraphics[width=5.5cm]{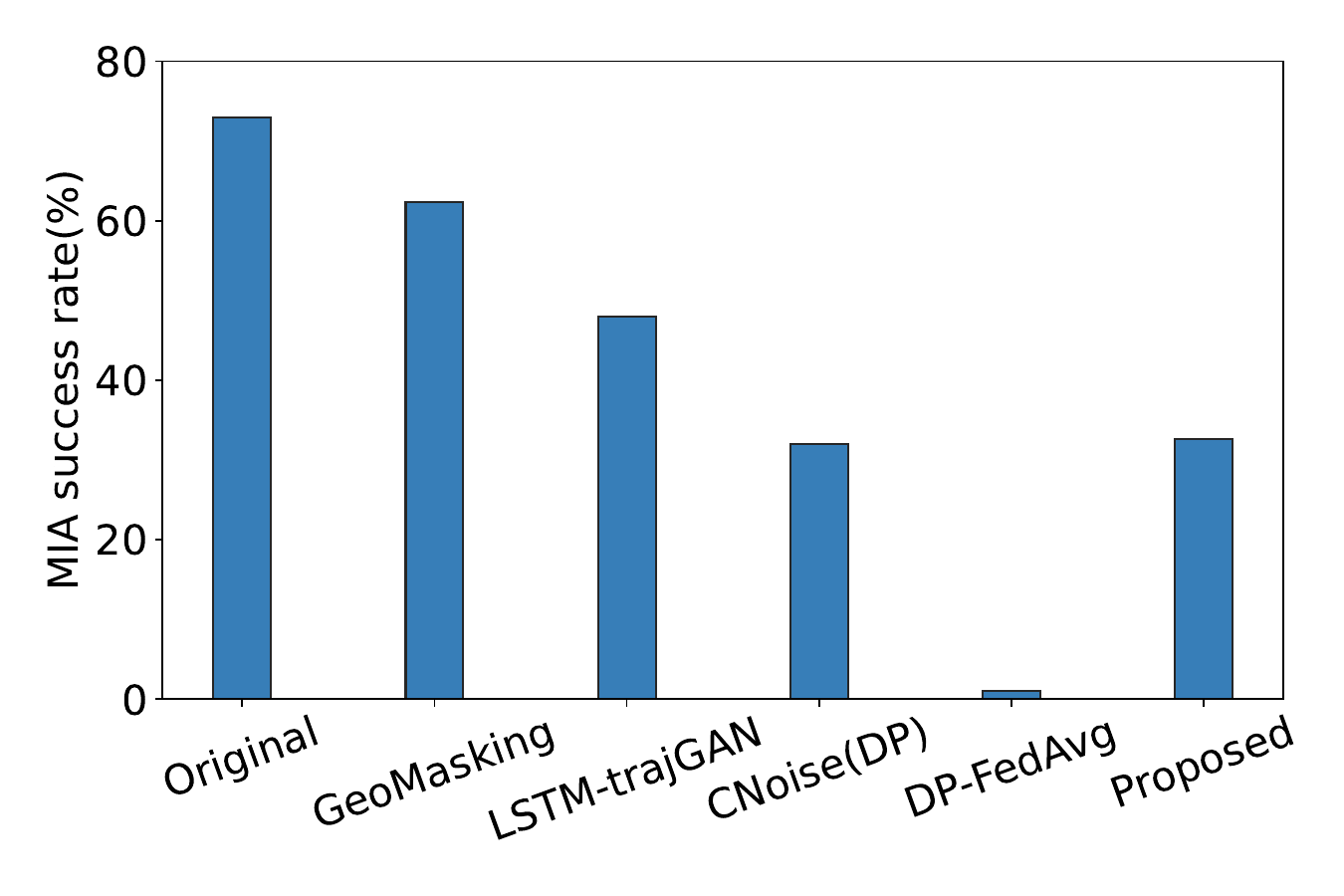}
    \caption{Success rate of Membership inference attack against taxi demand prediction.}
    \label{fig:MIA_success_rate}
  \end{minipage}
\end{figure*}

\color{black}
\subsection{Comparison with other FL methods}
\label{sec:eval_comparison_fl}
In this section, to measure the performance of \system, we compare the proposed system with the most relevant state-of-the-art techniques:
FedAvg\cite{pmlr-v54-mcmahan17a}, FedProx\cite{MLSYS2020_1f5fe839}, and FedProx with fine-tuning (FedProx-FT)\cite{wang2019federated}.

\textbf{FedAvg\cite{pmlr-v54-mcmahan17a}} is the most classic federated learning
method, which is the baseline of the training results.
\textbf{FedProx\cite{MLSYS2020_1f5fe839}} is widely used to make the training process efficient in non-IID datasets.FedProx uses a regulation parameter $\mu$ to prevent the local model from deviating greatly from the global model.
We fix the regulation parameter $\mu$ to $0.3$.
\textbf{FedProx-FT\cite{wang2019federated}} aims to build personalized models in non-IID settings by fine-tuning the global model trained in FedProx using local data.
\textbf{MOON\cite{li2021model}} is a federated learning approach that corrects the local updates in non-IID settings by maximizing the agreement of representation learned by the local model and the representation learned by the global model.

Figure \ref{fig:performance_comparison} displays the balanced accuracy of each method.
In our experiments, we have \textcolor{black}{five} local clients, and the values of other relevant parameters can be found in Table \ref{tab:config}.
All the reported performance metrics are averaged across clients for each distribution.
It is worth noting that the single model in the graph utilizes data from all clients, making its performance a reference. 
The local model in Figure\ref{fig:performance_comparison} represents the result of the prediction model trained with only local data.
In our observations, \system\ achieves the highest performance among the other state-of-the-art methods within decentralized settings.
This outcome highlights that \system\ excels by personalizing models for target regions while acquiring comprehensive knowledge through decentralized collaborative learning with similar regions.



\subsection{Comparison with other privacy preserved methods}
Our proposed system stands out for its security and privacy preservation at the architecture level because it does not require row data exchange, which ensures superior privacy protection compared to other methods.
Additionally, \system\ is designed to prevent overfitting of each facility's data, enhancing its security and robustness against machine learning model attacks, including membership inference attacks.
To validate the effectiveness and robustness of \system\ in terms of privacy-preserved architecture, we measured the performance of the machine learning model trained with privacy-preserved data and the robustness of these models against membership inference attacks.

For comparison purposes, we evaluate \system\ against the three state-of-the-art methods for data privacy protection: Geomasking \cite{gao2019exploring}, CNoise \cite{10.1145/2484838.2484846} and GAN-based\cite{ozeki2023balancing}.
Since we calculate taxi demand from trajectory data as mentioned in section\ref{sec:data_collection}, taxi demand calculated from privacy-preserved trajectory is also expected to be privacy preserved.
We also compared our proposed method with the state-of-the-art method for privacy protection of the machine learning model: DP-FedAvg\cite{mcmahan2017learning}.

\textbf{Geomasking\cite{gao2019exploring}} is a method that obscures trajectory data by applying spatio-temporal Gaussian noise (as outlined in \cite{gao2019exploring}).
\textbf{CNoise \cite{10.1145/2484838.2484846}} guarantees $\epsilon$-differential privacy by adding noise to location data via a Laplace distribution ($Lap(\mu=0, b=\frac{2\sqrt{2}S}{\epsilon})$). The privacy budget ($\epsilon$) is the desired level of privacy, which is achieved by choosing the appropriate noise level that fits the data set based on its extracted sensitivity metric ($S$).
\textbf{LSTM-trajGAN\cite{LSTM-trajGAN}} generates synthetic trajectory with LSTM and replaces the original dataset with the generated dataset.Since LSTM-trajGAN adopts a GAN-based approach for training the LSTM generator, the synthetic trajectory is expected to be realistic.
\textbf{DP-FedAvg\cite{mcmahan2017learning}} ensures privacy preservation by adding noise to global model weight based on differential privacy when the server aggregates model weights of each local model.

Figure \ref{fig:Comparison_architecture_result} displays the balanced accuracy of each method.
In our experiments, we have \textcolor{black}{five} local clients, and the values of other relevant parameters can be found in Table \ref{tab:config}.
In data privacy protection methods, we trained the taxi demand prediction model with privacy-preserved data.
From Figure\ref{fig:Comparison_architecture_result} \system outperformed the other state-of-the-art methods regarding prediction accuracy, which indicates our proposed system strikes the balance of privacy preservation and feasibility because our proposed method achieves the highest accuracy compared to other privacy-preserved methods while preserving privacy at the architecture level.


\subsection{Robustness against attack}
This section evaluates the proposed system's resilience against membership inference attacks (MIAs). 
MIAs aim to determine which data are included in the training dataset, potentially exposing users' spatio-temporal information.
To assess the effectiveness of \system against MIA (described in Section~\ref{sec:threat_mia}), we compared the success rate of inference attacks on our system's prediction model with models trained on synthetic data generated by geomasking, Differential Privacy (DP), and LSTM-trajGAN.

The results, presented in Figure \ref{fig:MIA_success_rate}, demonstrate that \system\ provides superior defense against MIAs, with a \textcolor{black}{40.4}\% better success rate compared to the original prediction model.
The original model is trained with raw data in Figure\ref{fig:MIA_success_rate}.
These findings underscore the exceptional capability of \system in preserving the privacy of users' Spatio-temporal information.
Moreover, \system provides the same level of privacy protection as CNoise, which is based on differential privacy, while ensuring the taxi demand prediction accuracy.
Since DP-FedAvg is a privacy-focused approach, DP-FedAvg offers a strong defense against MIAs.
However, DP-FedAvg compromises taxi demand prediction accuracy, resulting in lower performance than the local model, as illustrated in Figure\ref{fig:Comparison_architecture_result}.
These results indicate \system\ successfully balances privacy preservation with practical feasibility, ensuring robust privacy protection without sacrificing taxi demand prediction accuracy.

\section{Related work}
\label{sec:related_work}
\subsection{Taxi demand prediction}
Taxi demand prediction has recently garnered considerable attention owing to the abundance of large-scale spatiotemporal data that facilitates the training of deep neural networks, such as Convolutional Neural Networks (CNNs) and Long Short-Term Memory (LSTM) networks.

Recent studies have leveraged spatial and temporal characteristics to predict taxi demand more accurately.
For example, the author of \cite{Yao_Wu_Ke_Tang_Jia_Lu_Gong_Ye_Li_2018} employs a CNN to capture spatial features and an LSTM to capture temporal features, resulting in improved accuracy compared to methods that only consider semantic, spatial, or temporal information.
The author of \cite{9172100} recognizes the existence of spatio-temporal correlations between pick-up and drop-off locations and proposes a taxi demand prediction model using multitask learning, which predicts both pick-up and drop-off locations as interrelated tasks. This approach leads to more accurate prediction results.

Other studies have focused on accounting for the heterogeneity of taxi demand across regions.
The method proposed in \cite{9439926} clusters taxi demand data and trains region-specific models to predict demand, taking into account the unique distribution and temporal variability of demand in each region.
While these machine learning-based methods have shown promising results when applied to spatio-temporal data, they do not consider privacy implications.

\textit{In contrast, our proposed system evaluates the accuracy of taxi demand prediction while preserving privacy in distributed data situations. 
Our approach enables each facility to train a taxi demand prediction model independently, minimizing privacy risks through collaborative learning. 
This method ensures that raw data is not exchanged, addressing significant privacy concerns inherent in previous models.
}

\subsection{Privacy-preserving method for mobility data}
Spatio-temporal data is vital for training various data-driven approaches for various applications, such as taxi demand prediction.
However, collecting this data is costly and can reveal sensitive information, threatening user privacy.
Therefore, many works address preserving privacy for data and machine learning models.

Some studies \cite{4417165,10.1145/1869790.1869846} have focused on protecting user privacy by adding false information, known as dummy locations, to mobility data.
The method proposed in \cite{4417165} adds false location information to the trajectory in a random or rotating pattern.
Although this approach is effective when the attacker cannot judge the target trajectory as false, there is still a risk of an attacker detecting the authenticity of the created trajectory.
To create more realistic trajectories, the authors of \cite{10.1145/1869790.1869846} proposed a method to create dummy information by restricting user movements.
This method can achieve anonymity for a certain spatial area while also making the trajectory realistic. 
However, limitations exist, such as if the attacker knows about the user's lifestyle, they can still find out the user's interests or real location data \cite{HEMKUMAR20201291}.

Synthesis-based approaches for privacy protection for mobility data are also proposed, which replace the original dataset with the synthetic data \cite{LSTM-trajGAN, 4497446, zhu2023diffusion}.
K-anonymity is an important concept that ensures privacy preservation in the trajectory.
It is defined as the existence of $k$ or more users with the same characteristics, meaning that an attacker cannot reduce the number of candidates to $k$ or less, even if an attacker tries to identify a user based on specific characteristics.
To achieve k-anonymity for mobility data, a method proposed by the authors of \cite{4497446} adds uncertainty to location information.
To generate realistic and synthetic mobility data, DiffTraj\cite{zhu2023diffusion} generates a synthetic location dataset utilizing a diffusion model.
LSTM-trajGAN\cite{LSTM-trajGAN} also generates synthetic mobility data with LSTM and replaces the original dataset with the generated dataset.
Since LSTM-trajGAN adopts a GAN-based approach for training the LSTM generator, the synthetic trajectory is expected to be realistic.
However, these methods do not guarantee privacy preservation and utility for generated data.

To ensure privacy protection, differential privacy has emerged as the leading approach in anonymization and privacy preservation \cite{10.1145/2484838.2484846,10.1145/3423165}.
It is a mathematical framework that adds noise to data to guarantee a certain level of privacy protection.
Several methods have been proposed to implement differential privacy in trajectory data, such as CNoise and SDD \cite{10.1145/2484838.2484846}.
These methods add noise to mobility data to ensure differential privacy.
The added noise and the resulting privacy protection have been mathematically proven, making it a more robust approach than other methods.
Furthermore, to preserve the machine learning model directly from attacks against the machine learning model, such as membership inference attacks, some research proposed approaches that differential privacy applied to model training\cite{mcmahan2017learning, el2022differential, wei2020federated, adnan2022federated}.
This trend is particularly evident in the domain of federated learning, where the concept of imbuing differential privacy into client-specific model parameters has gained traction\cite{el2022differential, wei2020federated, adnan2022federated}. 
Such strategies are conceived to mitigate the inherent privacy risks confronting clients within federated learning ecosystems\cite{song2020analyzing, melis2019exploiting, NEURIPS2020_c4ede56b}. 
However, a major drawback of differential privacy is its limitations in balancing the performance-privacy trade-off.
To preserve privacy, it can negatively impact accuracy and performance. 


\textit{To address this challenge, \system incorporates a collaborative learning mechanism to train the model without publishing and exchanging data, which can preserve these sensitive data at the architecture level.
Additionally, since our system is designed to obtain comprehensive knowledge about taxi demand prediction, our system is expected to predict taxi demand more accurately, even in this privacy-preserved architecture.
}

\color{black}

\subsection{Machine Learning from Imbalanced Data}
\label{sec:learning_from_imbalanced_data}
As shown in Figure \ref{fig:imbalance_vis_hex}, taxi demand is absent in most areas because people use taxis in specific areas.
Additionally, since the scale and population of the target area that taxi companies focus on vary, the heterogeneity of label distribution may affect the practical performance of the prediction model in distributed learning.
Thus, the class imbalance problem is one of the fundamental problems of machine learning\cite{johnson2019survey, valverde2014100}.
For instance, in binary classification, the presence of majority and minority classes is assumed, with a specific imbalance ratio.
Since the standard classifier is optimized by a loss function that treats both classes equally, the minority class is almost ignored.
However, the minority class is more important to detect in usual cases that include our target research area.
To overcome this class imbalance problem, many researchers have addressed this problem\cite{tasci2022bias, alsaui2022resampling, singh2022credit}.
Approaches to solving class imbalance problems are mainly categorized into two: Data-level approach and algorithm-level approach.

The data-level approach aims to obtain balanced training datasets from imbalanced datasets utilizing undersampling or oversampling\cite{9533379, chawla2002smote, dablain2022deepsmote}.
Since undersampling methods risk removing data important for a model to learn the boundary, oversampling gains more attention than undersampling\cite{chawla2002smote, 8388338}.
To generate similar additional data even for multi-modal data, some data augmentation methods using generative models are proposed\cite{8388338} because it is difficult for the traditional statical approach such as SMOTE\cite{chawla2002smote} to generate additional high-quality multi-modal data.
The algorithm-level approach works directly during the training procedure of the classifier, disliking the previous data-level approach.
The most commonly addressed issue with the algorithm-level approach is loss function adaptation, such as 
Focal loss\cite{Lin_2017_ICCV}, cost sensitive learning\cite{9064578} and Mean False Error\cite{7727770}.
Mean False Error\cite{7727770} balances the weight of loss from minority and majority classes.
Focal loss\cite{Lin_2017_ICCV} reduces the impact of easy instances on the loss function.

The ensemble approach is the way to deal with imbalanced data by ensemble learning \cite{LIU201735}.
One of the most successful methods combines ensemble learning and undersampling\cite{DIEZPASTOR201596}.
These methods aim to classify accurately by applying ensemble learning to various balanced, stochastic, undersampled datasets.

While the data-level approach holds promise, the cost of data augmentation can be substantial, especially in scenarios with high-class imbalance ratios or multi-modal input data. 
Conversely, relying solely on loss function adaptation often proves ineffective when the classifier is a deep, complex neural network.
The amount of minority data is insufficient for obtaining high-quality latent representation.

\textit{To obtain meaningful knowledge even from imbalanced data, \system employs contrastive learning as self-supervised representation learning that does not rely on labels.
Furthermore, this approach makes \system robust to the heterogeneity of label distribution in distributed data situations.
}

\subsection{Distributed learning}
Federated learning (FL) is widely used when it is difficult to collect distributed data on a server due to privacy, security, or data migration costs because FL enables distributed clients to train a shared model collaboratively without exchanging local data\cite{HEGEDUS2021109}.
However, it is known that traditional FL cannot train efficiency when global data distribution and local data distribution differ, which is called non-IID\cite{MLSYS2020_1f5fe839}.
To build an effective model for global distribution in non-IID settings, many existing works address FL in non-IID settings\cite{MLSYS2020_1f5fe839, rothchild2020fetchsgd, li2021model, dong2021federated}.
FedProx \cite{MLSYS2020_1f5fe839} is designed to avoid the local model deviating greatly from the global model.

Although these works succeed in converging learning and improving the global model under non-IID data, the author of \cite{wang2019federated, huang2021personalized, dai2022dispfl} proposed personalization in FL because one global model cannot fit all clients.
One example is FedProx-FT\cite{wang2019federated}, which refines the global model from FedProx\cite{MLSYS2020_1f5fe839} using local data to create personalized models.
One of the similarity-based approaches is FedAMP\cite{huang2021personalized}, which keeps the personalized cloud model for each client trying to learn in each party without sharing each client's data.
MOON\cite{li2021model} is a federated learning approach that corrects the local updates by maximizing the agreement of representation learned by the current local model and the representation learned by the global model.
FedMoCo\cite{dong2021federated} uses federated learning only between similar nodes whose representations are similar.
This approach usually does not lead to model convergence and thus affects the system performance.
Additionally, none of the above-mentioned techniques has been applied to the field of taxi demand prediction.

\textit{In contrast, \system adopts a decentralized collaborative learning mechanism and neighbor-finding way based on weight similarity that ensures model convergence and thus saves in communication costs as demonstrated in \cite{9954190}.
}

\vspace{-0.3cm}

\section{Conclusion}
\label{sec:conclusion}
In this paper, we proposed a privacy-preserved taxi demand prediction system for distributed data.
Our proposed system aims to predict taxi demand accurately by utilizing contrastive learning and collaborative learning without exchanging raw data among facilities.
To address the data scarcity and heterogeneity of distributed taxi demand data, \system\ is designed to obtain comprehensive knowledge for taxi demand prediction and ensure model convergence.
To evaluate the effectiveness of \system, we collect real-world taxi demand data in five facilities and measure the accuracy of \system.
The results demonstrate that our approach achieves the highest accuracy in taxi demand prediction for distributed data compared to state-of-the-art approaches.
We also compared \system with other relevant privacy-preserved methods, such as differential privacy-based data protection in terms of the performance of the taxi demand prediction model and privacy protection.
Through the experiments, \system outperformed other relevant state-of-the-art privacy-preserved methods in terms of taxi demand prediction accuracy and achieves the same level of privacy protection with differential privacy-based methods.
This result demonstrated that our proposed system balances privacy protection and feasibility, which will benefit smart cities in the real world.


\bibliographystyle{ACM-Reference-Format}
\bibliography{sample-base}

\end{document}